\documentclass[conf]{new-aiaa}
\usepackage[utf8]{inputenc}
\usepackage[T1]{fontenc}
\usepackage{graphicx}
\usepackage{amsmath}
\usepackage[version=4]{mhchem}
\usepackage{siunitx}
\usepackage{longtable,tabularx}
\usepackage{ragged2e}
\usepackage{colortbl}
\usepackage{float}
\usepackage{multicol}
\usepackage{booktabs}

\title{Exploring the Boundaries: Thorough Software Testing for Safety-Critical Driving Scenarios Based on Kinematics in the Context of Autonomous Driving}

\author{Nico Schick\footnote{N. Schick, M. Sc. studied Applied Computer Sciences (M. Sc.) and Computer Engineering (B. Eng.) at the Esslingen University of Applied Sciences. e-mail: Nico.Schick@hs-esslingen.de}}

\begin{document}

\maketitle

\begin{abstract}
This scientific publication focuses on the efficient application of boundary value analysis in the testing of corner cases for kinematic-based safety-critical driving scenarios within the domain of autonomous driving. Corner cases, which represent infrequent and crucial situations, present notable obstacles to the reliability and safety of autonomous driving systems. This paper emphasizes the significance of employing boundary value analysis, a systematic technique for identifying critical boundaries and values, to achieve comprehensive testing coverage. By identifying and testing extreme and boundary conditions, such as minimum distances, this publication aims to improve the performance and robustness of autonomous driving systems in safety-critical scenarios. The insights and methodologies presented in this paper can serve as a guide for researchers, developers, and regulators in effectively addressing the challenges posed by corner cases and ensuring the reliability and safety of autonomous driving systems under real-world driving conditions.
\end{abstract}

\textbf{Scientific Question:} \textit{How can variety of kinematic-based safety-critical driving scenarios been tested based on boundary value analysis for autonomous driving?}


\section{SW Testing}
Software testing is a process of evaluating a software product or system to ensure that it meets the specified requirements and performs as expected \cite{SWTesting1}\cite{SWTesting2}. The objective of software testing is to identify defects or errors in the software and to ensure that the software is functioning as intended. There are several types of software testing \cite{SWTesting3}. Unit testing is the process of testing individual units or components of the software to ensure that they are functioning correctly \cite{SWTesting4}\cite{SWTesting5}. Integration testing is the process of testing the interaction between different units or components of the software to ensure that they work together as expected \cite{SWTesting6}. System testing is the process of testing the entire system as a whole to ensure that it meets the specified requirements and performs as expected \cite{SWTesting7}. Acceptance testing is the process of testing the software with the aim of determining whether or not it meets the requirements of the end-users or stakeholders \cite{SWTesting8}. 

Regression testing is the process of testing the software after changes or modifications have been made to ensure that the changes have not introduced any new defects or errors \cite{SWTesting9}. There are also several testing techniques that can be used in software testing \cite{SWTesting10}. Black box testing is a testing technique where the tester does not have access to the internal workings of the software and tests it based on the inputs and outputs \cite{SWTesting11}. White box testing is a testing technique where the tester has access to the internal workings of the software and tests it based on the code \cite{SWTesting12}. Grey box testing is a testing technique that combines both black box and white box testing. The tester has some knowledge of the internal workings of the software but not full access \cite{SWTesting13}. Literature on software testing covers topics such as test planning, Test Case design, test automation, test execution, and defect management \cite{SWTesting14}. 

\begin{figure}[H]
\centering
    \includegraphics[scale=0.48]{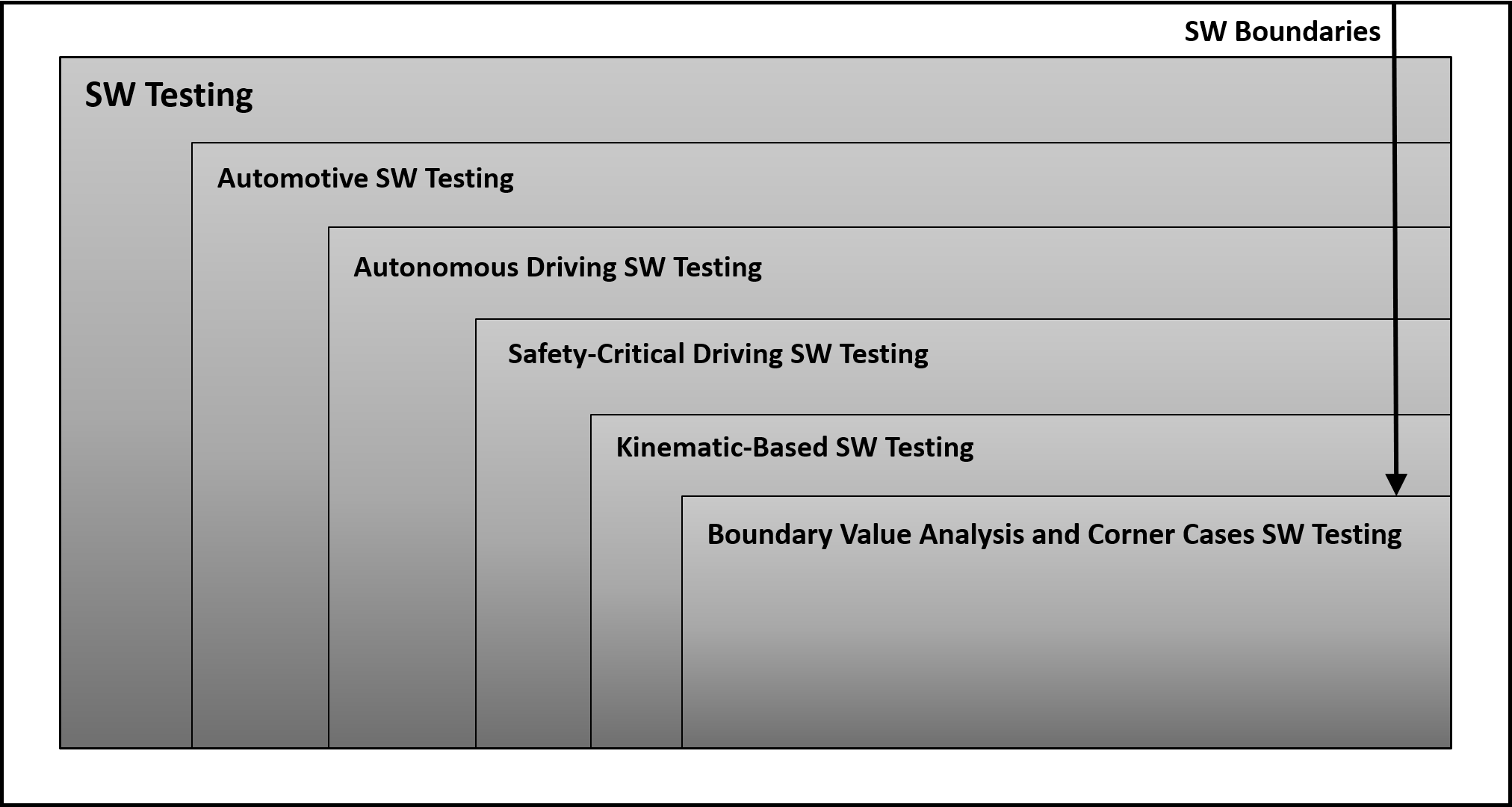}
    \caption{Different Levels of SW Testing}
    \label{Taxo2}
\end{figure}

In this publication, particular emphasis is given to corner case software tests based on kinematics. These tests are part of a broader spectrum of software testing levels that vary depending on the specific use case. Figure \ref{Taxo2} provides an overview of the different levels of detail, starting from the software testing level itself and extending up to the corner case test level. The subsequent chapters of the publication delve into each individual software test level, providing more detailed descriptions and insights.

\subsection{Automotive SW Testing}
Software testing in the automotive sector is a critical component of ensuring the safety, reliability, and quality of the final product \cite{SWTestingAutomotive1}. The automotive industry has specific requirements and regulations for software testing, as vehicles are complex systems with many interacting components that must be tested in a variety of scenarios \cite{SWTestingAutomotive2}. Testing in the automotive sector is often focused on functional safety, with the goal of preventing accidents caused by software failures \cite{SWTestingAutomotive3}. 

In addition to traditional testing methods, techniques such as model-based testing, hardware-in-the-loop testing, and software-in-the-loop testing are commonly used in the automotive industry \cite{SWTestingAutomotive4}. The importance of software testing in the automotive industry has led to the development of specialized testing frameworks and standards, such as ISO 26262, which provides guidelines for functional safety in automotive software \cite{SWTestingAutomotive5}. Automotive companies are also investing in new testing technologies, such as machine learning and artificial intelligence, to improve the efficiency and accuracy of testing processes \cite{SWTestingAutomotive6}. Overall, software testing plays a crucial role in the development and production of safe and reliable vehicles in the automotive sector \cite{SWTestingAutomotive7}.

\subsection{Autonomous Driving SW Testing}
Testing of autonomous driving software is essential for ensuring the safety and reliability of self-driving vehicles \cite{SWTestingAutonomesFahren1}. Due to the complexity of autonomous driving systems, testing approaches have evolved beyond traditional methods and now incorporate techniques such as simulation, virtual testing, and model-based testing \cite{SWTestingAutonomesFahren2} \cite{SWTestingAutonomesFahren3}. The testing of autonomous driving software involves not only functional testing, but also non-functional testing, including testing for performance, security, and safety \cite{SWTestingAutonomesFahren4} \cite{SWTestingAutonomesFahren5}. The development of autonomous driving software is subject to numerous regulations and standards, such as the Automated Driving System (ADS) Safety Framework developed by the National Highway Traffic Safety Administration (NHTSA) in the United States \cite{SWTestingAutonomesFahren6}. Testing of autonomous driving software also involves the use of large datasets and machine learning techniques to train the software to recognize and respond to different driving scenarios \cite{SWTestingAutonomesFahren7}. Overall, the testing of autonomous driving software is a critical component of ensuring the safety and reliability of self-driving vehicles \cite{SWTestingAutonomesFahren8}. 

\subsection{Safety-Critical Driving SW Testing}
Software testing is critical to ensuring the safety and reliability of autonomous driving systems, particularly in safety-critical scenarios where a failure could result in harm to humans or property \cite{SWTestingImpSCAuton1} \cite{SWTestingImpSCAuton2}. Testing software for safety-critical scenarios must be comprehensive and rigorous, covering a wide range of potential failure modes and edge cases that may not be encountered in normal driving scenarios \cite{SWTestingImpSCAuton3} \cite{SWTestingImpSCAuton4}. The use of model-based testing and simulation tools can help ensure that software is thoroughly tested and validated before it is deployed in the field \cite{SWTestingImpSCAuton5} \cite{SWTestingImpSCAuton6}. To ensure that software testing is effective in identifying potential safety-critical failures, it must be continuously updated and refined based on real-world data and feedback \cite{SWTestingImpSCAuton7} \cite{SWTestingImpSCAuton8}.

\subsection{Kinematic-Based SW Testing}
Kinematic-based Test Cases \cite{SWTesting14} for safety-critical driving scenarios are a type of Test Case used to evaluate the safety of a vehicle's motion based on kinematic principles such as velocity, acceleration, and steering. These Test Cases typically involve simulating the vehicle's motion in various scenarios, such as emergency braking or sudden lane changes, and analyzing whether the vehicle's motion satisfies safety constraints such as maintaining a safe distance from other vehicles and avoiding collisions. The use of kinematic-based Test Cases is important for ensuring the safety of autonomous vehicles and other advanced driver assistance systems, as it allows for the evaluation of the vehicle's motion under a wide range of driving scenarios.
\newpage
Kinematic-based Test Cases are a valuable tool for assessing the safety of autonomous vehicles in safety-critical driving scenarios. These Test Cases are designed to evaluate the vehicle's motion and assess its ability to respond to different situations, such as avoiding obstacles, maintaining lane position, or navigating through complex intersections. Various approaches can be used to create kinematic-based Test Cases, including simulation-based testing, real-world driving data analysis, and defining safety-critical scenarios and associated performance criteria. 

For example, Althoff et al. (2018) used a simulation-based approach to evaluate the safety of autonomous vehicles in complex urban environments \cite{KinematicBasedTC1}, while Sun et al. (2021) used naturalistic driving data to evaluate autonomous vehicle performance in different driving scenarios \cite{KinematicBasedTC2}. Several studies have focused on defining safety-critical scenarios and associated performance criteria to ensure comprehensive testing. Buss et al. (2020) proposed a framework that includes a set of safety-relevant scenarios and associated performance criteria covering a wide range of safety-critical situations, such as emergency braking, avoiding pedestrians, and merging onto a highway \cite{KinematicBasedTC3}. Similarly, Gressenbuch et al. (2021) proposed a comprehensive framework for evaluating the safety of automated driving systems, including the creation of kinematic-based Test Cases \cite{KinematicBasedTC4}. To ensure comprehensive testing, it is essential to consider various factors such as environmental conditions, road conditions, and the behavior of other road users. For example, Bujarbaruah et al. (2020) used kinematic-based Test Cases to evaluate the performance of autonomous vehicles in foggy conditions \cite{KinematicBasedTC5}, and Saad et al. (2021) evaluated the impact of different weather conditions on autonomous vehicle performance using kinematic-based Test Cases \cite{KinematicBasedTC6}. 

In conclusion, creating kinematic-based Test Cases requires a deep understanding of the safety-critical scenarios autonomous vehicles may encounter on the road, as well as a variety of testing approaches to evaluate vehicle performance in those scenarios. A comprehensive framework can ensure that researchers evaluate the safety of autonomous driving systems thoroughly and systematically.

\subsection{Boundary Value Analysis and Corner Cases SW Testing}
In the context of safety-critical driving scenarios for kinematic-based Test Cases, both Boundary Value Analysis (BVA) and corner case testing are important techniques to ensure the safety and reliability of the system. Boundary value analysis testing focuses on verifying the system's behavior at the limits of the acceptable range of input values, while corner case testing explores the system's behavior in unusual or extreme situations that are not normally encountered, such as emergency maneuvers or unexpected road conditions. The main difference between both is that boundary value analysis is more concerned with testing the system's response to inputs that are close to the limits of the acceptable range, while corner case testing is focused on exploring the behavior of the system in scenarios that are outside of the normal operating conditions. Both techniques are important in ensuring that the system is robust and reliable in all potential scenarios. \cite{BVACornerCases}

\newpage

\section{Kinematic-Based BVA for safety-critical driving scenario}
\begin{figure}[H]
\centering
    \includegraphics[scale=0.75]{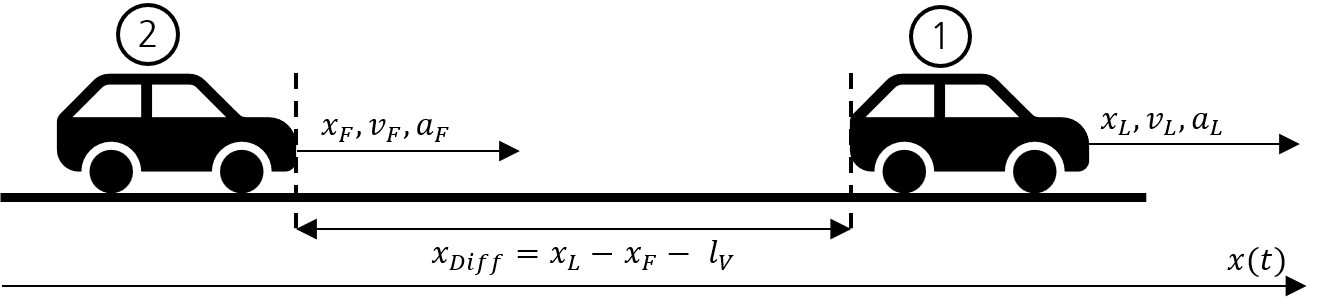}
    \caption{Car Following Drive (schema) \cite{pubS4}}
    \label{carfollowdrive}
\end{figure}

The follow-up drive scenario (visualized in Figure \ref{carfollowdrive}), where one vehicle follows another, is a widely recognized potential safety-critical driving situation. In this scenario, Vehicle 2 acts as the following vehicle, while Vehicle 1 serves as the leading vehicle. Both vehicles brake at certain times individually, with Vehicle 2 reacting to the braking of Vehicle 1. When considering this driving scenario based on the defined kinematic laws, four essential kinematic variables come into play: position (x), speed (v), acceleration (a), and time (t). Given the kinematic laws that define the driving scenario, the utilization of BVA becomes viable for examining the system's performance under extreme conditions.

\subsection{Difference Space Stopping (DSS)}
DSS is an appropriate safety indicator for the underlying use case \cite{pubS4} \cite{DSS}. In general, DSS is defined by the difference of space and stopping distance of two vehicles following each other. The space distance $a$ is calculated by the sum of the braking distance $x_{B,L}$ of the leading vehicle and the effective distance $d_V$ between leading and following vehicle. The stop distance $b$ is calculated by the sum of the brake reaction distance $x_{R,F}$ and the braking distance $x_{B,F}$ of the following vehicle. \cite{pubS4}
\begin{align}
DSS = a-b = \left(d_V + x_{B,L}\right) - \left(x_{R,F} + x_{B,F}\right)
\end{align}
In more detail, DSS depends on the position $x_L$ and velocity $v_L$ of the leading vehicle, the position $x_F$, velocity $v_F$ and reaction time $t_{B,R}$ of the following vehicle, the velocity difference of both $\Delta v$ as well as the maximum braking deceleration $a_{B,max}$ and vehicle's length $l_V$. \cite{pubS4}

The maximum braking deceleration $a_{B,max}$ in turn depends on gravitational acceleration $g$ and friction coefficient $\mu$ according to $a_{B,max} = g \, \mu$. By applying the maximum braking deceleration $a_{B,max}$, DSS can be interpreted as frozen position of both vehicles when the leading vehicle is braking at maximum suddenly, and therefore the following vehicle is also braking at maximum. \cite{pubS4}

In general, reaction times are dependent on psychologically and physiologically characteristics of the driver respectively of strength and timing of the actuating of the braking system. Drivers do not react perfectly in time and as reasonable in theory. Also the vehicle itself is a reacting system with delays and tolerances. \cite{pubS1} Autonomous cars are closing this gap even further. But nevertheless, there will be always a minimal gap at least between reacting and actuating of the vehicle braking system itself. So, taking real and more reliable driving values such as reacting times into account will lead to more reliable data sets for testing purposes of autonomous cars. 

Realistic reaction times $t_{B,R}$ turns DSS in a more generalized and reliable safety-indicator. Therefore, it increases also the potential to identify more realistic safety-critical driving scenario. For DSS, the reaction time $t_{B,R}$ can be modeled as a time-shifted Gamma distribution. \cite{pubS4} \cite{pubS1}

Mathematically, DSS can be defined based on both absolute-based ($DSS_{absolute}$) or relative-based ($DSS_{relative}$) as follows 
\begin{align}
DSS_{absolute} &= a-b=\left( \left( x_L-x_F-l_V \right) + \frac{v_{L}^2}{2 \, a_{B,max}} \right) - \left( v_F \, t_{B,R} + \frac{v_{F}^2}{2 \, a_{B,max}} \right) \\
DSS_{relative} &= a-b=\left( d_V + \frac{v_{L}^2}{2 \, a_{B,max}} \right) - \left( (v_L - \Delta v) \, t_{B,R} + \frac{(v_L - \Delta v)^2}{2 \, a_{B,max}} \right)
\end{align}
and with substitution for relative representation as follows
\begin{align}
d_V &:= x_L-x_F-l_V \\
\Delta v &= v_L - v_F \leftrightarrow v_F := v_L - \Delta v
\end{align}

\subsubsection{Effective Distance}
\begin{figure}[ht]
\centering
    \includegraphics[scale=0.75]{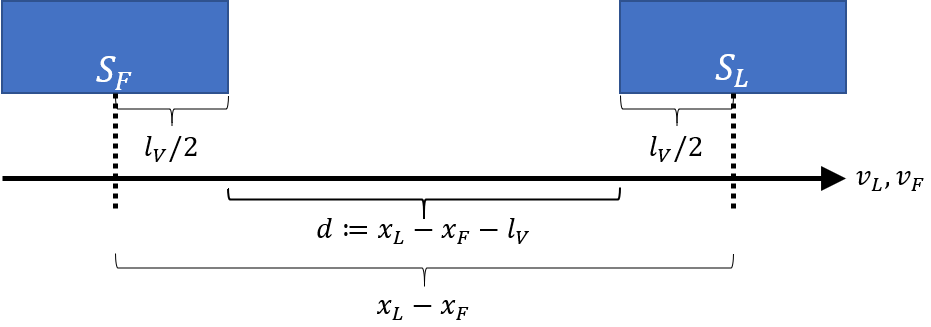}
    \caption{Effective Distance (Longitudinal Driving) \cite{pubS4}}
    \label{fig:vehicle-length}
\end{figure}

The significance of vehicle length for metrics of this nature is depicted in Figure \ref{fig:vehicle-length}. The distance between two vehicles is influenced by the positions of both vehicles and the length of the following vehicle, represented as $l_V$. Therefore, the safety-related distance between the vehicles, referred to as the effective distance, can be defined as $d := x_L - x_F - l_V$. It is important to note that this calculation assumes a uniform mass distribution for both vehicles, with the center of gravity denoted as $S$ and positioned at the midpoint. \cite{pubS4}

Achieving comprehensive coverage of potential driving scenarios is of paramount importance for safety indicator metrics like DSS. Evaluating this coverage efficiently can be accomplished by considering a range of possible driving scenarios. One effective approach involves utilizing matrices that focus on speed, acceleration, or deceleration combinations for both vehicles, particularly when they are dependent on each other. These matrices offer a structured representation of the scenarios and facilitate a systematic analysis of the safety indicator's effectiveness in capturing various critical situations. (e.g. follow-up drive). \cite{pubS4} The following matrices (Table \ref{vMatrix} and Table \ref{aMatrix}) provide an example of such an assessment \cite{pubS4}:
			\begin{table}[H]
			\centering
			\begin{minipage}{0.45\textwidth}		
	\begin{tabular}{|l|l|l|l|}
		\hline
		& \textbf{$v_L < 0$} & \textbf{$v_L = 0$} & \textbf{$v_L > 0$} \\ \hline
		\textbf{$v_F < 0$} & 0 & 0 & 0 \\ \hline
		\textbf{$v_F = 0$} & 0 & 0 & 0 \\ \hline
		\textbf{$v_F > 0$} & 0 & 1 & 1 \\ \hline
	\end{tabular}
			\caption{\label{vMatrix} $v$ Matrix (Safety-relevant) \cite{pubS4}}
			\end{minipage}\hfill
			\begin{minipage}{0.45\textwidth}
	\begin{tabular}{|l|l|l|l|}
		\hline
		& \textbf{$a_L < 0$} & \textbf{$a_L = 0$} & \textbf{$a_L > 0$} \\ \hline
		\textbf{$a_F < 0$} & 1 & 0 & 0 \\ \hline
		\textbf{$a_F = 0$} & 1 & 0 & 0 \\ \hline
		\textbf{$a_F > 0$} & 0 & 0 & 0 \\ \hline
	\end{tabular}
			\caption{\label{aMatrix} $a$ Matrix (Safety-relevant) \cite{pubS4}}
			\end{minipage}
			\end{table}
The matrices presented in this study define the ranges of speed and acceleration for both vehicles, with rows and columns representing these values. A cell value of 0 indicates a non-safety-critical driving scenario, whereas a cell value of 1 signifies a safety-critical driving scenario. It is important to highlight that the matrices, along with their corresponding cell values, pertain to potential rear-end collisions and are specifically associated with DSS. In essence, there are four specific combinations of speed and acceleration that yield a cell value of 1. Consequently, achieving comprehensive coverage for a safety indicator metric, such as DSS, necessitates including precisely these four combinations in the testing process. \cite{pubS4} Hence, DSS covers following potential rear-end collision in practise \cite{pubS4}: 
\begin{itemize}
\item Vehicle in front moves forward or comes to a standstill while the vehicle behind moves forward.
\item Vehicle in front brakes while the vehicle behind either also brakes or continues to drive at constant speed.
\end{itemize}

\subsection{Derivation of Test Cases}
The number of DSS-related Test Cases (TC) based on BVA depends on whether the absolute-based or relative-based approach of DSS is considered. In general, both the absolute-based and relative-based approaches depend on different parameters. Those parameters can be differentiated into variable and constant parameters. In this regard, parameter considering to be constant are maximum braking deceleration $a_{B,max}$ respectively gravitational acceleration $g$ and friction coefficient $\mu$ as well as vehicle's length $l_V$ (considering uniform vehicle sizes). The number of variable parameters is denoted by $n_{apa}$ for $DSS_{absolute}$ and denoted by $n_{apr}$ for $DSS_{relative}$. In addition, variable parameters can be differentiated into reasonable and unreasonable ones.  Unreasonable parameters are characterized by the fact that they are not explicitly necessary for the calculation of the safety indicator such as DSS. With respect to $DSS_{relative}$, the sizes $x_L$ and $x_F$ as well as $v_L$ and $v_F$ are of type unreasonable parameters because they can be also represented by $d_V$ and $\Delta v$ ($DSS_{absolute} \rightarrow DSS_{relative}$). All other parameters are reasonably variable parameters. 
\begin{align}
DSS_{absolute,f} &= f(x_L,x_F,v_L,v_F,t_{B,R}) \\
DSS_{relative,f} &= f(d_V,\Delta v,t_{B,R})
\end{align}
So, it results in 5 reasonable variable parameters $(x_L,x_F,v_L,v_F,t_{B,R})$ corresponding to $DSS_{absolute}$ denoted by $n_{rpa}=5$ as well as 3 reasonable variable parameters $(d_V,\Delta v,t_{B,R})$ corresponding to $DSS_{relative}$ denoted by $n_{rpr}=3$.

Whether the absolute ($DSS_{absolute}$) or relative approach ($DSS_{relative}$) is used depends on the use case as well as which data set is available. However, depending on data availability, both approaches can be transformed into each other. 

The fewer test cases used to achieve test coverage, the greater the time savings, transparency, and effectiveness. With respect to targeting a low number of TCs for achieving test coverage, considering of $DSS_{relative}$ is more suitable than $DSS_{absolute}$ due to the lower number of reasonably variable parameters. For this reason, only the relative-based DSS approach ($DSS_{relative}$) is considered further in more detail.

Applying BVA on safety-critical driving scenarios turns into splitting of TCs for non-safety-critical and safety-critical perspective. In addition, a permutational approach can be used to identify the individual TCs. More specifically, for each reasonably variable parameter there exist one parameter value at least for both a non-safety-critical case and safety-critical case as well. It results in a safety-critical variation factor denoted by $n_{crit,var}=2$ to be constant. With respect to DSS and specific reasonably variable parameter, one parameter value can be chosen to be greater or equal ($DSS_{relative} \geq 0$) or smaller ($DSS_{relative} < 0$) than DSS threshold. Based on $n_{rpr}=3$ reasonably variable parameter for $DSS_{relative}$, it results in $N_{rpr} = n_{rpr} \, n_{crit,var} = 3 \cdot 2 = 6$ TCs for $DSS_{relative}$ perspective in total. 

\begin{figure}[H]
\centering
    \includegraphics[scale=0.6]{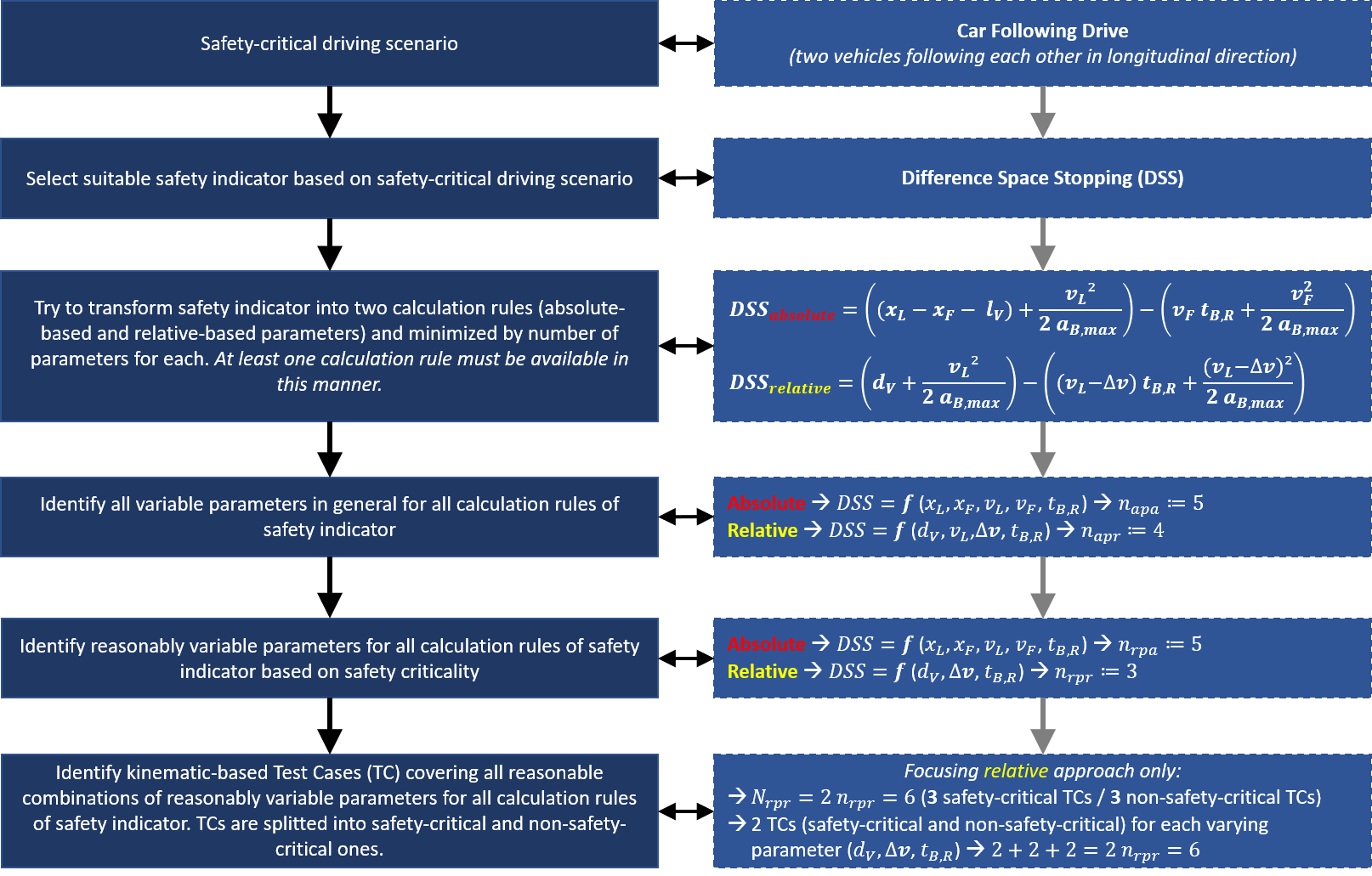}
    \caption{Derivation of Kinematic-Based TCs based on Safety Indicator}
    \label{dxTC}
\end{figure}

Overall, the derivation of kinematic-based TCs for safety-critical driving scenarios based on safety indicator like DSS is shown in Figure \ref{dxTC}.

\newpage

For $DSS_{relative}$, those $N_{rpr}=6$ kinematic-based TCs are examined further. Considering related $n_{rpr}=3$ reasonably variable parameters ($d_v, \Delta v, t_{B,R}$) correspond to meaningful statements regarding DSS perspective. In this regard, Table \ref{tab:TC} shows the definitions of the 6 TCs on higher level (qualitative). Each TC has an unique identifier named by TC.$x$ ($x = 1,2,\ldots,N_{rpr}$). The TCs are also clearly splitted into safety-critical and non-safety-critical ones: TC.1, TC.3 and TC.5 are safety-critical (SC), whereas TC.2, TC.4 and TC.6 are non-safety-critical (NSC).  
\begin{table}[H]
\begin{tabular}{|p{2.1cm}|p{1.9cm}|p{1.9cm}|p{8.9cm}|}
\hline
 \textbf{\# TC} & \textbf{Criticality} & \textbf{Parameter} & \textbf{Description} \\ \hline
 \midrule
 \textbf{TC.1} & SC & $d_v$ & Effective distance is too small \\ \hline
 \textbf{TC.2} & NSC & $d_v$ & Effective distance is large enough \\ \hline
 \textbf{TC.3} & SC & $\Delta v$ & Velocity difference is too high \\ \hline
 \textbf{TC.4} & NSC & $\Delta v$ & Velocity difference is small enough \\ \hline
 \textbf{TC.5} & SC & $t_{B,R}$ & Reaction time is too high \\ \hline
 \textbf{TC.6} & NSC & $t_{B,R}$ & Reaction time is small enough \\ \hline
\end{tabular}
\caption{\label{tab:TC} Test Cases for Relative-Based DSS (Qualitative)}
\end{table}

Test cases must be testable. That is why Table \ref{tab:TC2} shows definitions of respective TCs on lower level (quantitative). The TCs itself are arranged column-wise, whereas parameters are arranged row-wise. The first three rows correspond to the $n_{rpr}=3$ reasonably variable parameter data for $d_v$, $\Delta v$ and $t_{B,R}$ leading into safety-critical and non-safety-critical driving scenarios based on $DSS_{relative}$ and Table $\ref{tab:TC}$. The parameters $d_v = 42.56 \pm 0.01$ m, $\Delta v = -5.\overline{5}6 \pm 0.0028$ m/s $(20$ km/h $\pm 0.01$ km/h$)$ and $t_{B,R} = 0.7 \pm 0.0003$ s were varied in such a way that the safety rating according to DSS differs based on absolute accuracy of 0.01 m (1 cm).

\begin{table}[H]
\centering
\begin{tabular}{|p{2.1cm}|p{1.9cm}|p{1.9cm}|p{1.9cm}|p{1.9cm}|p{1.9cm}|p{1.9cm}|}
\hline
\textbf{Parameter} & \textbf{TC.1} & \textbf{TC.2} & \textbf{TC.3} & \textbf{TC.4} & \textbf{TC.5} & \textbf{TC.6} \\ \hline
\midrule
$d_v \; [m]$ & $42.55$ & $42.57$ & $42.56$ & $42.56$ & $42.56$ & $42.56$ \\ \hline
$\Delta v \; [m/s]$ & $-5.\overline{5}6$ & $-5.\overline{5}6$ & $-5.5583$ & $-5.5528$ & $-5.\overline{5}6$ & $-5.\overline{5}6$ \\ \hline
$t_{B,R} \; [s]$ & $0.7$ & $0.7$ & $0.7$ & $0.7$ & $0.7003$ & $0.6997$ \\ \hline
\midrule
$a \; [m]$ & $86.25$ & $86.27$ & $86.26$ & $86.26$ & $86.26$ & $86.26$ \\ \hline
$b \; [m]$ & $86.26$ & $86.26$ & $86.27$ & $86.25$ & $86.27$ & $86.25$ \\ \hline
$DSS \; [m]$ & $-0.01$ & $0.01$ & $-0.01$ & $0.01$ & $-0.01$ & $0.01$ \\ \hline
\midrule
Criticality & SC & NSC & SC & NSC & SC & NSC \\ \hline
\end{tabular}
\caption{\label{tab:TC2} Test Cases for Relative-Based DSS (Quantitative)}
\end{table}

\newpage 

\section{Conclusion}
The handling of safety-critical driving scenarios is imperative for autonomous vehicles. Consequently, achieving a broad test coverage of autonomous driving behaviors within these safety-critical scenarios holds significant importance. To efficiently test safety-critical driving scenarios, methods such as boundary value analysis can be employed, specifically focusing on corner cases in software testing. In this publication, we provide an illustration of this approach by applying it to a safety-critical Car Following Drive, utilizing Difference Space Stopping (DSS) as a safety indicator. Furthermore, we present a methodology for identifying a minimal set of test cases that offer extensive coverage for the specific use case, considering the chosen safety-critical driving scenario and safety indicator.

\end{document}